\documentclass[letterpaper, 10 pt, conference]{ieeeconf}  % Comment this line out if you need a4paper

\IEEEoverridecommandlockouts                              % This command is only needed if 
\usepackage{cite}
\usepackage{amsmath,amssymb,amsfonts}
\usepackage{algorithmic}
\usepackage{graphicx}
\usepackage{textcomp}
\usepackage{xcolor}
\usepackage{cuted}
\usepackage{capt-of}
\usepackage{float}
\usepackage{booktabs}
\usepackage{threeparttable}
\usepackage{pgf}
\usepackage{tikz}

\usepackage[ruled,linesnumbered]{algorithm2e}

\usetikzlibrary{arrows, decorations.pathmorphing, backgrounds, positioning, fit, petri, automata}

\title{\LARGE \bf
Pointly-supervised 3D Scene Parsing with Viewpoint Bottleneck
}

\author{Liyi Luo$^{1,2}$, Beiwen Tian$^{1}$, Hao Zhao$^{3}$ and Guyue Zhou$^{1}$% <-this % stops a space
\thanks{$^{1}$ Institute for AI Industry Research (AIR), Tsinghua University, China {tbw18@mails.tsinghua.edu.cn, zhouguyue@air.tsinghua.edu.cn}}%
\thanks{$^{2}$ McGill University, Canada liyi.luo@mail.mcgill.ca}%
\thanks{$^{3}$Intel Labs China, Peking University, China  
        zhao-hao@pku.edu.cn, hao.zhao@intel.com}}%

\begin{document}

\maketitle
\thispagestyle{empty}
\pagestyle{empty}

%%%%%%%%%%%%%%%%%%%%%%%%%%%%%%%%%%%%%%%%%%%%%%%%%%%%%%%%%%%%%%%%%%%%%%%%%%%%%%%%
\begin{abstract}

Semantic understanding of 3D point clouds is important for various robotics applications. Given that point-wise semantic annotation is expensive, in this paper, we address the challenge of learning models with \textit{extremely sparse labels}. The core problem is how to leverage numerous unlabeled points. To this end, we propose a self-supervised 3D representation learning framework named viewpoint bottleneck. It optimizes a mutual-information based objective, which is applied on point clouds under different viewpoints. A principled analysis shows that viewpoint bottleneck leads to an elegant surrogate loss function that is suitable for large-scale point cloud data. Compared with former arts based upon contrastive learning, viewpoint bottleneck operates on the feature dimension instead of the sample dimension. This paradigm shift has several advantages: It is easy to implement and tune, does not need negative samples and performs better on our goal down-streaming task. We evaluate our method on the public benchmark ScanNet, under the pointly-supervised setting. We achieve the best quantitative results among comparable solutions. Meanwhile we provide an extensive qualitative inspection on various challenging scenes. They demonstrate that our models can produce fairly good scene parsing results for robotics applications. Our code, data and models will be made public.
\end{abstract}

%%%%%%%%%%%%%%%%%%%%%%%%%%%%%%%%%%%%%%%%%%%%%%%%%%%%%%%%%%%%%%%%%%%%%%%%%%%%%%%%
\section{INTRODUCTION}

\begin{figure}[t]
\centerline{\includegraphics[width=0.48\textwidth]{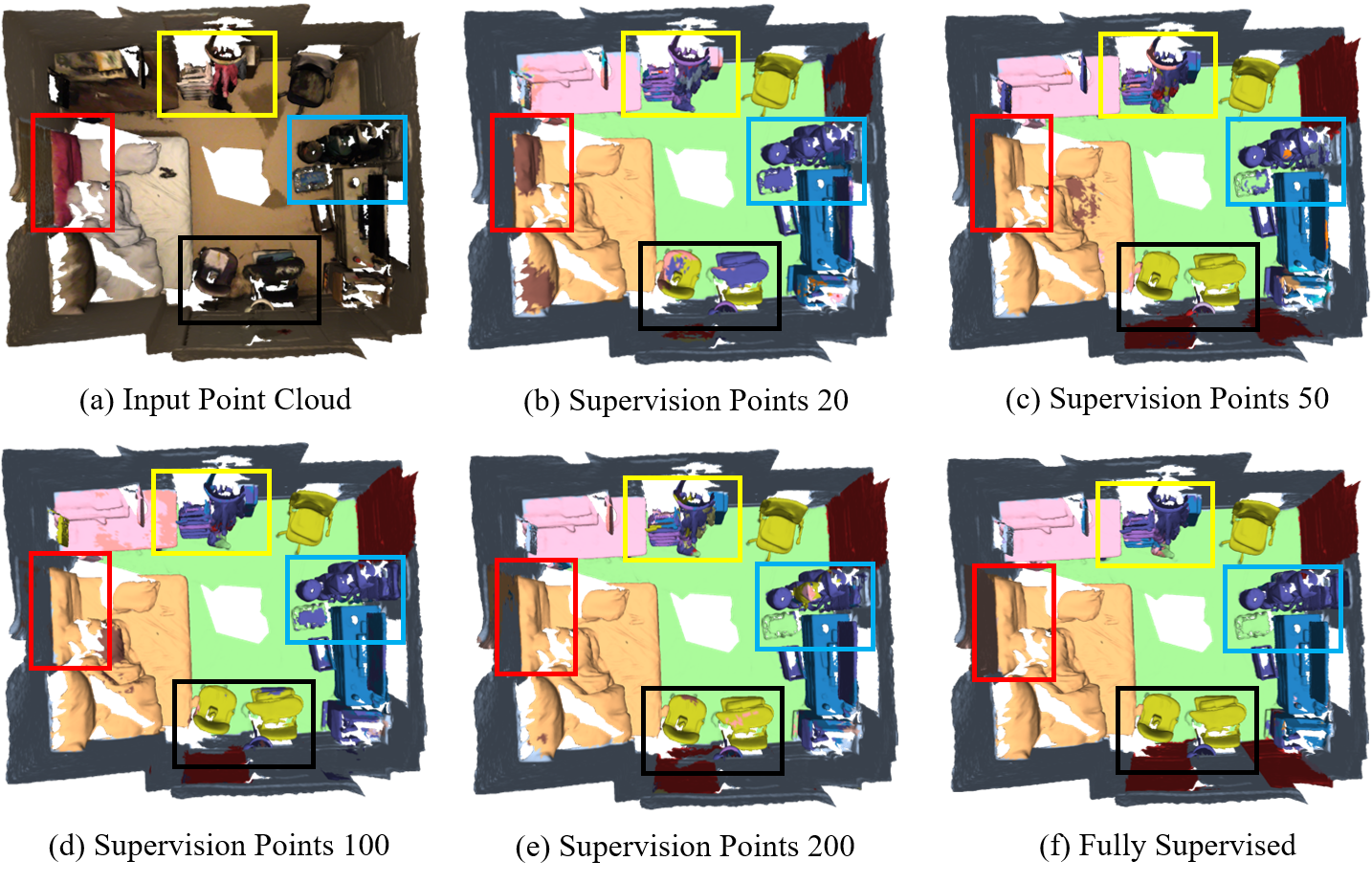}}
\caption{We address the challenging problem of learning 3D scene parsing models using only semantic annotations on limited points (20 to 200 points per scene). (a) shows a point cloud of an indoor scene. (b)-(e) demonstrate 3D semantic understanding results using 20 to 200 point annotations per scene. (f) shows the result generated by a fully supervised model. Except for highlighted regions in boxes, most points are correctly segmented, showing the effectiveness of our viewpoint bottleneck method.}
\label{fig:abstract_imag}
\end{figure}

We study the problem of dense semantic parsing of 3D point clouds. As illustrated by Fig.~\ref{fig:abstract_imag}-(a), a bedroom can be reconstructed by TSDF-based fusion methods like \cite{b1} \cite{b2}. Then taking it as input, a 3D scene parsing network generates a semantic label for each point in it, as demonstrated by Fig.~\ref{fig:abstract_imag}-(f). This is a core environment sensing capability for intelligent robots. Knowing the semantic class of the 3D world lays the foundation for decision making. However, training such a model needs point-wise annotation, which is very expensive and time-consuming to obtain. As such, we address the challenge of learning such models using an extremely small subset of annotations. For example, Fig.~\ref{fig:abstract_imag}-(e) shows the parsing result generated by a model trained using 200 points per scene, which is almost as good as Fig.~\ref{fig:abstract_imag}-(f). Since a scene in the ScanNet database \cite{b3} typically contains more than 50000 points, a subset of 200 points only amounts to 0.4\% data usage.

The key to effective pointly-supervised learning is leveraging numerous unlabelled points (e.g., the other 99.6\% data in the aforementioned case). If we can learn meaningful representations from them without the usage of semantic annotation, a subsequent pointly-supervised fine-tuning is expected to give good results. This is usually referred to as self-supervised representation learning (SSRL). While exciting progress has been achieved in 2D SSRL \cite{b4}\cite{b5}\cite{b6}, 3D SSRL in point clouds \cite{b7}\cite{b8} is still an under-explored emerging topic. Many open problems exist and challenge the effectiveness of existing 3D SSRL methods:

(1) Former 2D SSRL methods treat each image as a sample. While 3D SSRL for dense semantic parsing naturally requires us to treat each point as a sample. This fact leads to an extremely large sample set.

\begin{figure}[t]
\centerline{\includegraphics[width=0.48\textwidth]{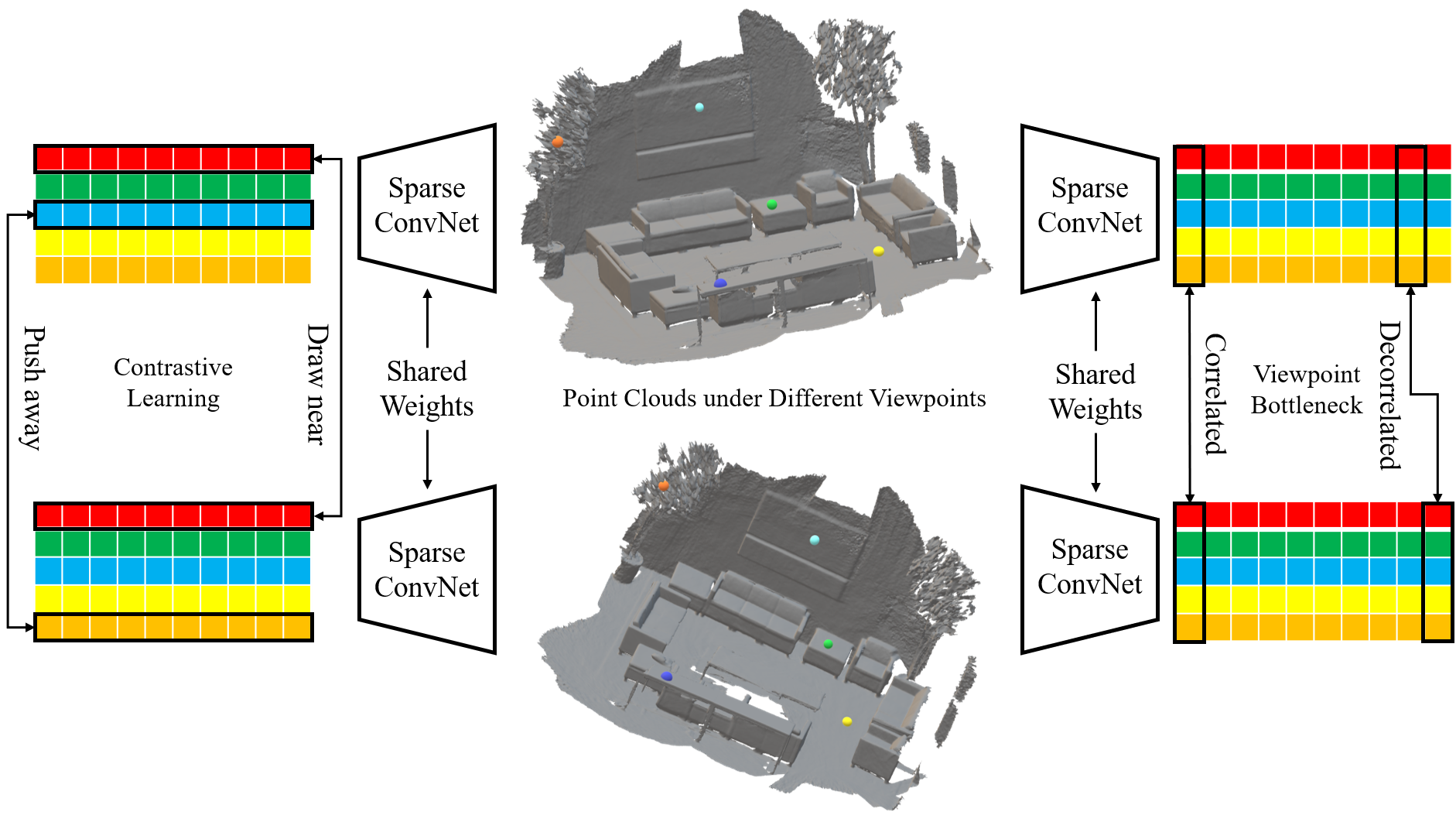}}
\caption{Five colored dots demonstrate five point samples, under random geometric transformations. Vectors of corresponding colors demonstrate features generated by a shared Sparse ConvNet. Conventional contrastive learning methods are shown in the left. Same samples under different viewpoints (red) are drawn near while different samples (blue and orange) are pushed away, in the embedding space. Viewpoint bottleneck is shown in the right. It operates on the feature dimension instead of the sample dimension. The corresponding channel is correlated while different channels are decorrelated.}
\label{fig:teaser}
\end{figure}

(2) Former contrastive learning methods operate on the sample dimension. As conceptually shown in Fig.~\ref{fig:teaser}-left, embeddings for the same sample under different viewpoints are drawn near, and embeddings for different samples are pushed away. Selecting good sample pairs is an unclear and difficult problem, especially when the 3D sample set is large.

(3) Former contrastive learning methods are troubled with degenerated solutions thus requires sophisticated techniques to break the symmetry like weight averaging with momentum \cite{b6}. Complex implementation details make them hard to tune.

Very recently, several works \cite{b10}\cite{b11}\cite{b12} propose to switch representation learning from the sample dimension to the feature dimension. Inspired by them, we propose a new 3D representation learning method called viewpoint bottleneck. It starts with an objective defined with the mutual information between features and samples. A principled analysis under mild distribution assumptions shows that an efficient surrogate loss implementation can be used to optimize the objective. The surrogate loss is conceptually shown in Fig.~\ref{fig:teaser}-right. In one word, it maximizes the correlation between corresponding feature channels while decorrelates different feature channels. This paradigm shift brings some advantages over former 3D SSRL methods based on contrastive learning, like PointContrast \cite{b7} or Contrastive Scene Context \cite{b8}. There is no need to select sample pairs or design sophisticated symmetry breaking techniques, leading to a simple implementation. This simpleness does not come with a prize of performance drop. By contrast, viewpoint bottleneck is more effective than the state-of-the-art (SOTA) method Contrastive Scene Context \cite{b8}. When evaluated on our goal downstream task pointly-supervised scene parsing, it consistently out-performs \cite{b8} in all inspected settings. To summarize, we make following contributions:

\begin{itemize}
\item[$\bullet$] We propose a new self-supervised 3D representation learning framework named viewpoint bottleneck, which is free of typical drawbacks of contrastive learning. 
\item[$\bullet$] We provide a principled analysis to motivate viewpoint bottleneck, which is absent in former 3D SSRL methods based upon contrastive learning.
\item[$\bullet$] We demonstrate the effectiveness of learned representations for pointly-supervised scene parsing. We evaluate on the public benchmark ScanNet and achieve consistently better results than existing SOTA 3D SSRL solutions. A public implementation is provided.
\end{itemize}

\section{Related Work}

\subsection{Self-supervised Representation Learning}

SSRL is a special form of unsupervised learning, which aims to learn meaningful representations of the input data without relying on annotations. The common ground of these methods is to learn representations that are invariant to distortions. There are several different ways to achieve this generic principle. For example, early attempts use different surrogate tasks whose labels can be naturally generated, like pixel value inpainting \cite{b21}\cite{b23}, cross-channel feature regression \cite{b22}, or rotation prediction \cite{b24}.  Recently, contrastive learning \cite{b27}, which was firstly proposed for metric learning, has drawn significant attention and witnessed great success for SSRL. However, it is now clear that a trivial contrastive learning formulation suffers from degeneration. As such, many variants have been proposed to address the issue like weight averaging \cite{b6}\cite{b28} or stop gradient \cite{b29} among others \cite{b14}\cite{b26}. 

Compared to its 2D counterpart, 3D SSRL is more urgent because of the difficulty of annotating 3D data. There are already some pioneering works that address the 3D SSRL problem borrowing ideas of contrastive learning, like PointContrast\cite{b7} and Contrastive Scene Contexts\cite{b8}. The former proposes a PointInfoNCE loss and verifies its effectiveness on a diverse set of scene understanding tasks. However, it ignores the spatial context around local points, which limits its transferability to complex downstream tasks. The latter introduces a loss function that contrasts features aggregated in local partitions. It demonstrates the possibility of using extremely few annotations to obtain good performance on a series of weakly-supervised settings. \cite{b9}\cite{b13} propose to use multi-modal pairs and tuples as elements for RGB-D constrastive learning and scene understanding. \cite{b35} develops a sophisticated spatio-temperal contrastive learning framework for point clouds.

To address the common issues of contrastive learning, a recent SSRL method \cite{b10} proposes an objective function that measures the cross-correlation matrix between the representations of two distorted versions of a sample. It naturally avoids representation collapse and sample selection since no negative samples are required. \cite{b11}\cite{b12} enrich this new scheme with a variance term and a principled kernel-based statistical measure. Inspired by them, we propose the viewpoint bottleneck principle, which is the first non-contrastive 3D SSRL method.

\subsection{Weakly-supervised Scene Parsing}

State-of-the-art scene parsing approaches generally depend on expensive sample-wise annotation to obtain good performance. To this end, many weakly-supervised solutions are proposed. \cite{b15} exploits deep metric learning to regularize the embedding, while collecting pseudo labels online by expansion. \cite{b25} identifies a statistical phenomenon called uncertainty mixture to harvest labels in an unsupervised manner. BoxSup\cite{b16} generates pseudo labels by evaluating the overlap between box supervision and bottom-up region proposals. STC\cite{b17} treats salient regions as the initial pseudo labels for image-level semantic classes. \cite{b18} proposes a method that exploits class activation maps (CAMs) to generate positive and negative affinity labels for pixel pairs. For 3D parsing, \cite{b19} generates pseudo point-wise labels by applying CAM on sub-point-clouds. \cite{b20} transfers 2D image annotations to 3D space to produce labels in point cloud. In this paper, we follow the path of 3D SSRL introduced by \cite{b7}\cite{b8} and propose a formulation free of sample selection.

\section{Method}

In this section, we describe the procedure of learning pointly-supervised scene parsing models with viewpoint bottleneck. As shown in Fig.\ref{fig:overview}, our approach is consisted of two stages: (1) a 3D SSRL stage using the newly proposed viewpoint bottleneck loss, and (2) fine-tuning the pretrained model for scene parsing with point supervision.

\subsection{Overview}
A fully-supervised 3D scene parsing dataset is denoted as $\{P_i, L_i\}, i\in\{1,2,...,N\}$, in which $P_i (M\times 6)$ is the point cloud represented by the concatenation of 3D coordinates and colors and $L_i (M\times 1)$ is the point-wise semantic label set for $P_i$. A weakly-supervised 3D scene parsing dataset is depicted as $\{P_i, S_i\}$, where $S_i$ is an annotation-efficient version of $L_i$. In $S_i$, only few points have been labeled and other points' labels are set to 
\emph{None}. Firstly, we train a pre-trained model $\rm VB(*,\Theta)$ only using $P_i$, as shown in Fig.\ref{fig:overview}-(A). Secondly, we fine-tune the pre-trained model $\rm VB(*,\Theta)$ with the weakly-supervised scene parsing dataset $\{P_i, S_i\}$. A new prediction head is added as illustrated by the light blue block in Fig.\ref{fig:overview}-(B). The network parameterized by $\Psi$ predicts the probability $Q(y_{ijs}|P_{ij}
;\Psi)$, in which $j$ indexes points in $P_i$ and $s$ indexes semantic categories. We train it with the Softmax cross-entropy loss as below:

\begin{equation}
L_{\rm ce}=-\frac{1}{NM}\sum_{i=1}^{N}\sum_{j=1}^{M}{-\log{Q(y_{ij\bar{s}}|P_{ij};\Psi)}}\label{eqce}
\end{equation}

Note here $\bar{s}$ is the ground truth label for $P_{ij}$. Finally, we run the trained scene parsing model, as in Fig.\ref{fig:overview}-(C).

\subsection{Self-supervised Pre-training}
Here we describe our self-supervised learning stage. 

\begin{figure*}[t]
\centerline{\includegraphics[width=0.88\textwidth]{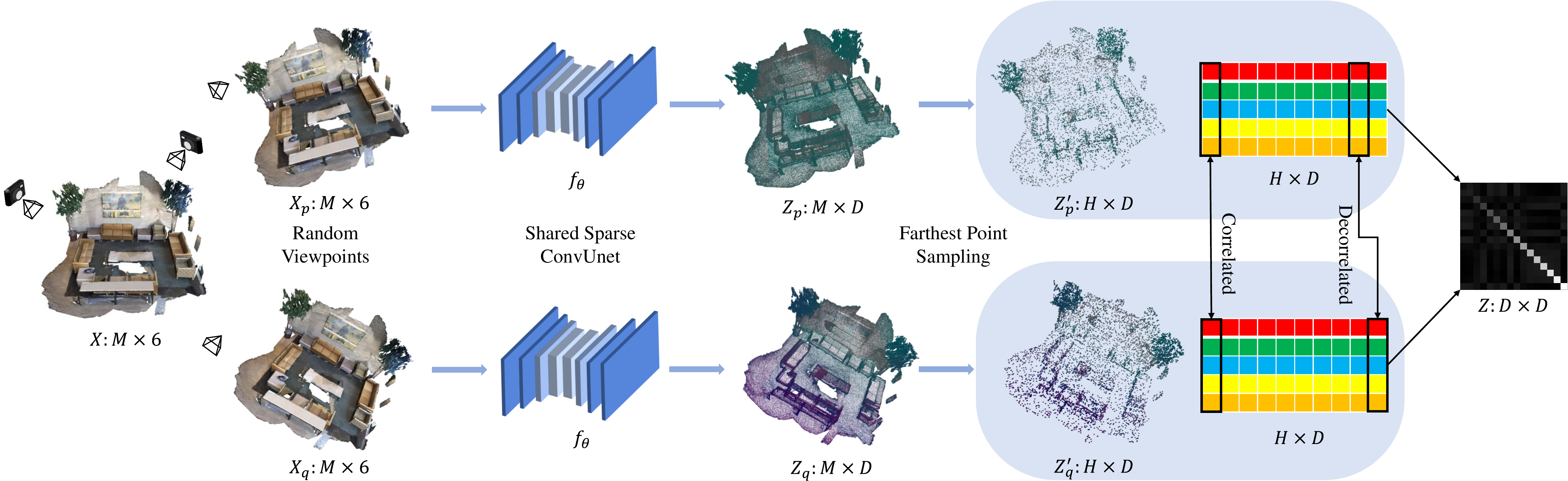}}
\caption{Illustration of viewpoint bottleneck. $X$ is a point cloud represented by the concatenation of 3D coordinates and colors. After two random geometric transformations, we obtain its two augmentations $X_p$ and $X_q$. They are sent to a shared Sparse ConvUnet $f_\theta$, forming two high-dimensional feature sets $Z_p$ and $Z_q$. To keep computation tractable, we apply farthest point sampling on them to get down-sampled $Z_p'$ and $Z_q'$. Finally, the viewpoint bottleneck is imposed on the cross-correlation matrix between $Z_p'$ and $Z_q'$, which is denoted as $Z$.}
\label{fig:pre-train}
\end{figure*}

Firstly, a point cloud $X={\{p_i\}}$ is augmented into two viewpoints by random geometric transformations. Transformed 3D point clouds are denoted by $X_p=\{p_i'\}$ and $X_q=\{p_i''\}$ respectively. As Fig.~\ref{fig:pre-train} shows, these two randomly transformed point clouds are fed into the same sparse convolutional network $f_\theta$ to obtain two high-dimensional representations $Z_p = f_\theta(X_p)$ and $Z_q = f_\theta(X_q)$. 

Next, we sample the representations by Farthest Point Sampling (FPS) from $M\times D$ to $H\times D$, where $D$ is the dimension of the representations and $M,H$ are the numbers of points before and after sampling. FPS starts with a random point as the first proposal candidate, and iteratively selects the farthest point from the already selected points until $H$ candidates are selected. We set $H\ll M$ to keep computation tractable, while FPS guarantees that the sampled subset is a reasonable abstraction of the original point cloud.

After FPS, the cross-correlation matrix $\mathcal{Z} (D\times D)$ is computed on the batch dimension, from two sampled representations $Z_{p}'$ and $Z_{q}'$. Finally, the network is trained with the proposed viewpoint bottleneck loss as detailed later. %We adopt the Minkowski Convolutional Neural Network \cite{b34} as the backbone and only use 3D coordinates and color information of point clouds as the network's inputs.

\subsection{Viewpoint Bottleneck: Analysis}

\begin{figure}[t]
\centerline{\includegraphics[width=0.48\textwidth]{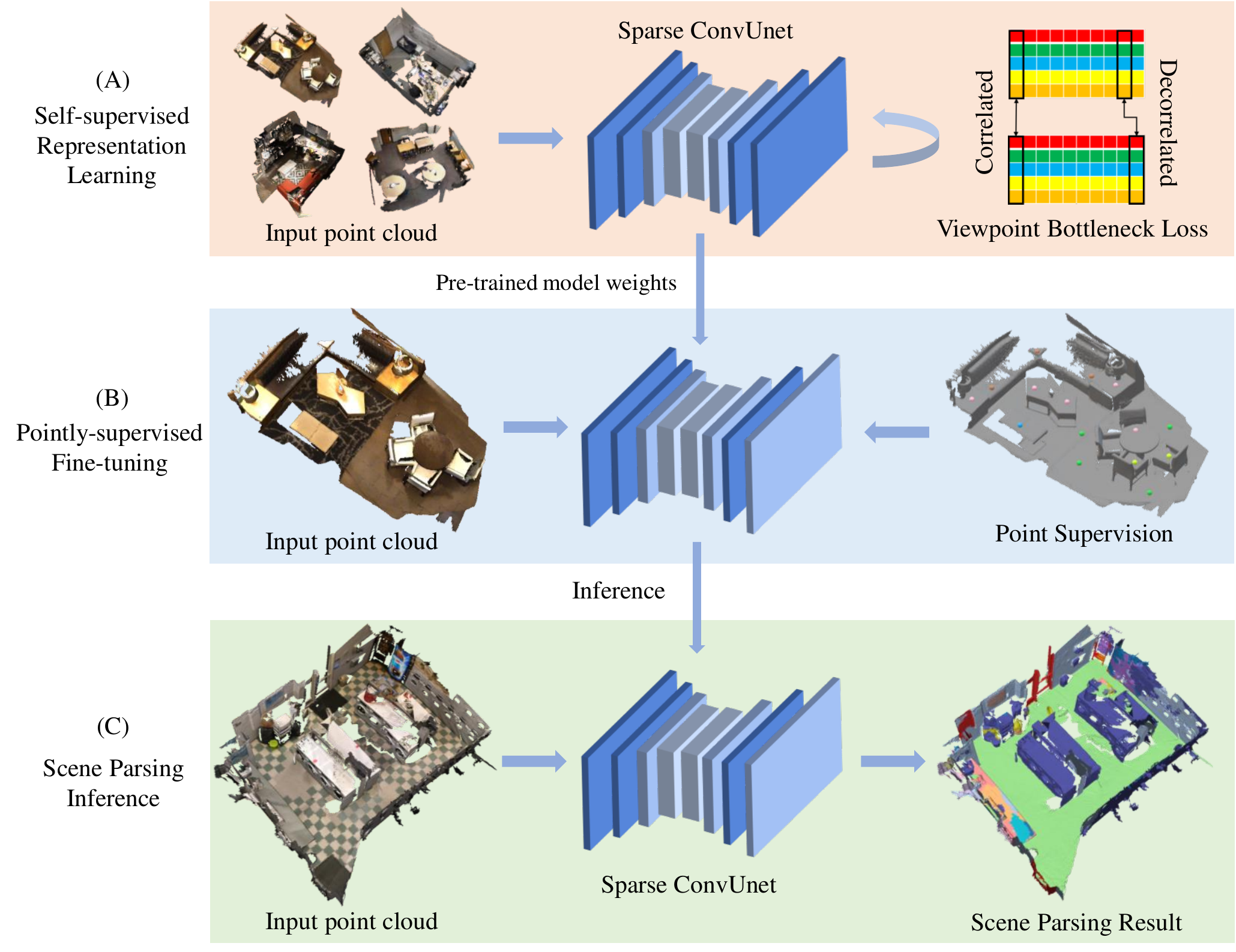}}
\caption{The overall pipeline of our method. In the first stage (A), we pretrain the backbone network with a self-supervised learning objective named viewpoint bottleneck. In the second stage (B), we fine-tune the weights from (A) for scene parsing, while only using a very small number of annotated points. Finally, we illustrate an inference stage (C).}
\label{fig:overview}
\end{figure}

\begin{figure}[t]
\centerline{\includegraphics[width=0.48\textwidth]{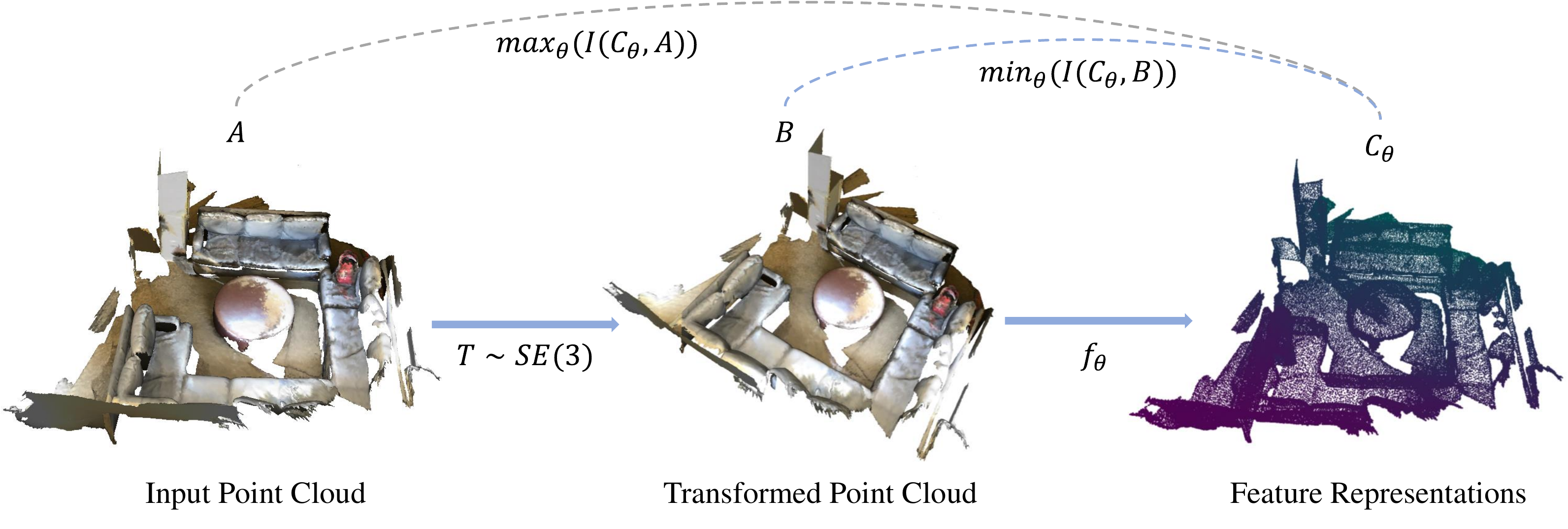}}
\caption{A conceptual illustration of the viewpoint bottleneck principle, in which $A$, $B$ and $C_\theta$ are random vectors, representing the original input point clouds, randomly transformed point clouds and high-dimensional representations generated by $f_\theta$ respectively. We aim to minimize the mutual information between $B$ and $C_\theta$ and maximize the mutual informaiton between $A$ and $C_\theta$. As such, there exists a bottleneck at $B$.}
\label{fig:VBP}
\end{figure}

Inspired by Barlow Twins \cite{b10}, we derive our viewpoint bottleneck objective from an information theory perspective \cite{b30}\cite{b31}. As shown in Fig.\ref{fig:VBP}, the information bottleneck principle assumes that meaningful representations should reserve as much information as possible about the input, while affected by random viewpoint transformations as little as possible. Formally the objective is stated as:
\begin{equation}
{\rm VB}_\theta\triangleq{I(C_\theta,B)-\beta I(C_\theta,A)}\label{eq:IB}
\end{equation}
where $I(\cdot,\cdot)$ is mutual information between two random vectors. Here $A$, $B$, and $C_\theta$ correspond to input point clouds, randomly transformed point clouds and learnt representations respectively. $\beta$ is a balancing weight. By definition, the mutual information can be calculated as $I\left(C_\theta,B\right)=H\left(C_\theta\right)-H\left(C_\theta\middle|B\right)$ (similarly for $I\left(C_\theta,A\right)$). $H\left(C_\theta\right)$ is the entropy of $C_\theta$ while $H\left(C_\theta\middle|B\right)$ is the conditional entropy of $C_\theta$ given $B$. As the function $f_\theta$ is deterministic, $H(C_\theta|B)$ cancels to 0. We rewrite the Eq.\ref{eq:IB} as:
\begin{equation}
{\rm VB}_\theta=H\left(C_\theta\right)-\beta\left[H\left(C_\theta\right)-H\left(C_\theta\middle|A\right)\right]\label{eq:entropy}
\end{equation} 

%representation and input point cloud, $I(C_\theta,B)$ is the mutual information between representation and transformed point cloud, and $\beta$ is a positive scalar parameter, trading off protection of original information and being invariant to transformation.

%There is a classical identity for mutual information:
%\begin{equation}
%\label{eq:N}
%\end{equation}
%\begin{equation}
%I\left(C_\theta,A\right)=H\left(C_\theta\right)-H\left(C_\theta\middle|A\right)\label{eq:M}
%\end{equation}
%where $H(.)$ denotes entropy. $H(C_\theta|B)$ is the entropy of the representation conditioned on a particular viewpoint scene. Because  Meanwhile, using Eq.\ref{eq:N} and Eq.\ref{eq:M},  
By dividing Eq.\ref{eq:entropy} with $\beta$, we have:
\begin{equation}
{\rm VB}_\theta=H\left(C_\theta\middle|A\right)+\frac{1-\beta}{\beta}H\left(C_\theta\right)\label{eq:ibp}
\end{equation}

However, the optimization of ${\rm VB}_\theta$ becomes a challenge as evaluating the entropy of a generic random vector is intractable. We derive a surrogate loss as below:

We assume the representation $C_\theta$ is distributed as a Gaussian: $C_\theta = (c_{\rm \theta i})_{i=1}^D \sim\mathcal{N}_D(\mu,\Sigma_{C_\theta})$. $\Sigma_{C_\theta}$ can be decomposed into $\Sigma_{C_\theta}\ =\ E^TE$, and $Y=E^{-1}({\vec{C_\theta}-\vec{\mu}})\sim\mathcal{N}\left(\vec{0},I\right)$. In the following derivation, $|\cdot|$ means taking determinant. The probability density function of $C_\theta$ is: 
\begin{equation}
\begin{split}
   &p\left(c_{\rm \theta 1},c_{\rm \theta 2},\cdots,c_{\rm \theta D}\right)\\
   &=\frac{1}{\left(2\pi\right)^\frac{n}{2}\left|\Sigma_{C_\theta}\right|^\frac{1}{2}}e^{-\frac{1}{2}\left[\left({\vec{C_\theta}-\vec{\mu}}\right)^T\Sigma_{C_\theta}^{-1}\left(\vec{C_\theta}-\vec{\mu}\right)\right]} 
\end{split}
\end{equation}
Then according to \cite{b33}:
% \begin{equation}
% \begin{split}
% H\left(C_\theta\right)&=-\int{p\left(c_\theta\right)\log{p\left(c_\theta\right)dc_\theta}}\\
% &=-\int{p\left(c_\theta\right)\log{\frac{e^{-\frac{1}{2}\left[\left({\vec{C_\theta}-\vec{\mu}}\right)^T\Sigma_{C_\theta}^{-1}\left(\vec{C_\theta}-\vec{\mu}\right)\right]}}{\left(2\pi\right)^\frac{n}{2}\left|\Sigma_{C_\theta}\right|^\frac{1}{2}}}dc_\theta}\\
% &{=-\int{p\left(c_\theta\right)\left[\log{\frac{1}{\left(2\pi\right)^\frac{n}{2}\left|\Sigma_{C_\theta}\right|^\frac{1}{2}}}+\log{e^{-\frac{1}{2}\left[\left({\vec{C_\theta}-\vec{\mu}}\right)^T\Sigma_{C_\theta}^{-1}\left(\vec{C_\theta}-\vec{\mu}\right)\right]}}\right]dc_\theta}}\\
% &=\log{\left(2\pi\right)^\frac{n}{2}\left|\Sigma_{C_\theta}\right|^\frac{1}{2}}-\int{p\left(c_\theta\right)\log{e^{-\frac{1}{2}\left[\left({\vec{C_\theta}-\vec{\mu}}\right)^T\Sigma_{C_\theta}^{-1}\left(\vec{C_\theta}-\vec{\mu}\right)\right]}}dc_\theta}\\
% &=\log{\left(2\pi\right)^\frac{n}{2}\left|\Sigma_{C_\theta}\right|^\frac{1}{2}}-\int{p\left(y\right)\log{e^{-\frac{1}{2}{\vec{Y}}^T\vec{Y}}}dy}\\
% &=\log{\left(2\pi\right)^\frac{n}{2}\left|\Sigma_{C_\theta}\right|^\frac{1}{2}}-\sum_{i=1}^{n}\log{e^{-\frac{1}{2}\left|y_i^2\right|}}\\
% &=\log{\left(2\pi\right)^\frac{n}{2}\left|\Sigma_{C_\theta}\right|^\frac{1}{2}}+\log{e^\frac{n}{2}}\\
% &=\frac{1}{2}\log{{(2 \pi e)}^n}\left|\Sigma_{C_\theta}\right|\label{eq}
% \end{split}
% \end{equation}
\begin{equation}
\begin{split}
H\left(C_\theta\right)&=-\int{p\left(c_\theta\right)\log{p\left(c_\theta\right)dc_\theta}}\\
&=-\int{p\left(c_\theta\right)\log{\frac{e^{-\frac{1}{2}\left({\vec{C_\theta}-\vec{\mu}}\right)^T\Sigma_{C_\theta}^{-1}\left(\vec{C_\theta}-\vec{\mu}\right)}}{\left(2\pi\right)^\frac{D}{2}\left|\Sigma_{C_\theta}\right|^\frac{1}{2}}}dc_\theta}\\
&=-\int{p\left(c_\theta\right)\log{\frac{1}{\left(2\pi\right)^\frac{D}{2}\left|\Sigma_{C_\theta}\right|^\frac{1}{2}}}}dc_\theta\\
&\quad-{\int{p\left(c_\theta\right)\log{e^{-\frac{1}{2}\left({\vec{C_\theta}-\vec{\mu}}\right)^T\left(E^{-1}\right)^{T}E^{-1}\left(\vec{C_\theta}-\vec{\mu}\right)}}dc_\theta}}\\
&=\log{\left(2\pi\right)^\frac{D}{2}\left|\Sigma_{C_\theta}\right|^\frac{1}{2}}-\int{p\left(Y\right)\log{e^{-\frac{1}{2}Y^TY}}dY}\\
&=\log{\left(2\pi\right)^\frac{D}{2}\left|\Sigma_{C_\theta}\right|^\frac{1}{2}}-\sum_{i=1}^{D}\log{e^{-\frac{1}{2}\left|y_i^2\right|}}\\
&=\log{\left(2\pi\right)^\frac{D}{2}\left|\Sigma_{C_\theta}\right|^\frac{1}{2}}+\log{e^\frac{D}{2}}\\
&=\frac{1}{2}\log{{(2 \pi e)}^D}\left|\Sigma_{C_\theta}\right|\label{eq}
\end{split}
\end{equation}
% \begin{equation}
% H\left(C_\theta\right)=\frac{1}{2}\log{{(2 \pi e)}^n}\left|\Sigma_{C_\theta}\right|\label{eq}
% \end{equation}
\begin{algorithm}
    \caption{3D SSRL with Viewpoint Bottleneck}\label{algorithm:1}
    \KwIn{
    \begin{itemize}
            \item[1] $\mathbb{X}$: a set of 3D point clouds; 
            \item[2] $\mathcal{T}$: the distribution of geometric transformations;
            \item[3] $f_\theta$: sparse ConvUnet parameterized by $\theta$;
            \item[4] $K$: the number of optimization steps;
            % \item[5] $N$: batch size.
            \item[5] $\lambda$: the control parameter in VB loss
    \end{itemize}}
    \KwOut{feature network $f_\theta$}
    \For{$k=1$ \KwTo $K$}{
        % $\mathbb{B} \longleftarrow \{X_i \in \mathbb{X}\}_{i=1}^N$
        
        % \For{$i=1$ \KwTo $N$}{
        \tcc{sample transformations}
            $T_{p},T_{q} \sim \mathcal{T}$
        
        \tcc{take an input point cloud}
            $X \in \mathbb{X}$
            
        \tcc{apply transformations}
            $X_{p}=T_{p}(X), \  X_{q}=T_{q}(X)$
            
        \tcc{extract features} 
            $Z_{p}={f}_{\theta_k}(X_{p}), \  Z_{q}={f}_{\theta_k}(X_{q})$
            
        \tcc{farthest point sampling}    
            $Z_{p}' = {\rm FPS}(Z_{p}), \ Z_{q}' = {\rm FPS}(Z_{q})$
            
        \tcc{calculate cross-correlation}
            $\mathcal{Z} = (Z_{p}')^T Z_{q}'$
        
        \tcc{calculate VB loss}
            % $\mathcal{L}_{\rm VB} =  \sum_{i}\left(1-{\mathcal{Z}}_{ii}\right)^2+\lambda\sum_{i}\sum_{j\neq i}{\mathcal{Z}}_{ij}^2\label{eq:vb}$ \bold{todo check}
            % $\mathcal{L}_{\rm VB}=\|\mathcal{Z}-\mathcal{I}_{\lambda}\|_2 \label{eq:vb}$
            $\mathcal{L}_{\rm VB}= \|\Gamma_{\lambda}(\mathcal{Z})-\mathcal{I}\|_F$
        % }
        
        \tcc{optimize network parameters}
        
        $\theta_{k+1}={\rm optim}(\theta_{k},\mathcal{L}_{\rm VB})$
        
        }
\end{algorithm}

\begin{equation}
\begin{split}
H\left(C_\theta\middle|A\right) &=\sum_{a} p\left(a\right)H\left(C_\theta\middle|A=a\right)\\
&=\frac{1}{2}\mathbb{E}_A\log(2 \pi e)^{n}|\Sigma_{C_{\theta}|A}|\label{eq}
\end{split}
\end{equation}

As such the viewpoint bottleneck objective Eq.\ref{eq:ibp} is turned into a practical surrogate loss function as:
\begin{equation}
{\rm VB}_\theta=\mathbb{E}_A\log{\left|\Sigma_{C_\theta|A}\right|}+\frac{1-\beta}{\beta}\log{\left|\Sigma_{C_\theta}\right|}\label{eq:ibf}
\end{equation}

Finally, we make several more simplifications: 1) Since Eq.\ref{eq:ibf} is only meaningful when $\beta > 1$, we use a positive constant $\lambda$ to replace $\frac{\beta-1}{\beta}$. 
2) In the second term of the Eq.\ref{eq:ibf} we use the Frobenius norm of $\Sigma_{C_\theta}$ as the metric.
3) Recall that the first term of Eq.\ref{eq:ibf} is derived from the conditonal entropy of the representation $C_\theta$ given $A$. Minimizing it is equivalent to maximizing the diagonal elements of $\mathcal{Z}$ (the aforementioned $D \times D$ cross-correlation matrix). %If the representation completely depend on the original scene, the $H\left(C_\theta\middle|A\right)$ cancels to 0 (minimizing the information of the representation contains about the viewpoint). The first term of Eq.\ref{eq:vb} maximizes the alignment between representations of two viewpoint scenes. Therefore, they have the same global optimum.

\begin{figure*}[t]
\centerline{\includegraphics[width=1\textwidth]{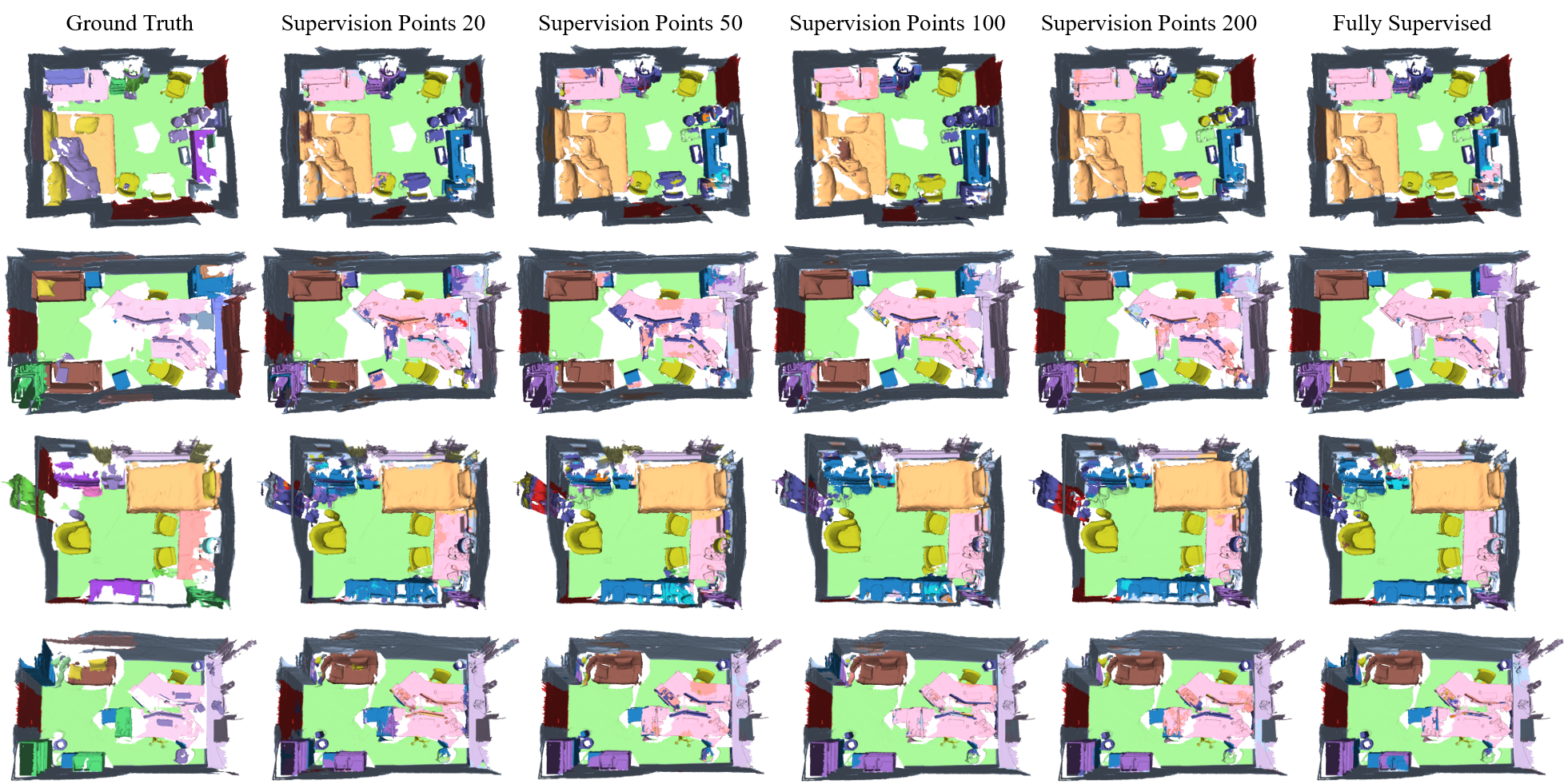}}
\caption{Qualitative visualizations for point-supervised 3D scene parsing. The first column is ground truth and the last is generated by a fully-supervsied model. The middle columns are generated by models pretrained by viewpoint bottleneck and supervised by sparse point annotations. }
\label{fig:all_results}
\end{figure*}

\begin{table}
\centering
\caption{Pointly-supervised 3D Parsing results on the ScanNet validation set (mIOU:\%)}
\begin{tabular}{lcccc}
    \toprule
    Points & $200$ & $100$ & $50$ & $20$\\
    \midrule
    Baseline &  64.3 & 60.6  & 55.4 & 45.9\\
    CSC \cite{b8} & 68.2 (+3.9) & 65.9 (+5.3)  & 60.5 (+5.1) & 55.5 (+9.6) \\
    VB (256) & 68.4 (+4.1) & 66.5 (+5.9)  & 63.3 (+7.9) & 56.2(+10.3) \\
    VB (512) & \textbf{68.5 (+4.2)} & \textbf{66.8 (+6.2)}  & 63.6 (+8.2) & \textbf{57.0 (+11.1)} \\
    VB (1024) & 68.4 (+4.1) & 66.5 (+5.9)  & \textbf{63.7 (+8.3)} & 56.3 (+10.4) \\
    \bottomrule
\end{tabular}
\label{tbl:table2}
\end{table}

\begin{table}[t]
\centering
\caption{Results reported by ScanNet online test server (mIOU:\%)}
\begin{tabular}{lcccc}
    \toprule
    Points & 200 & 100 & 50 & 20\\
    \midrule
    Viewpoint\_BN\_LA\_AIR (ours) & \textbf{66.9} & \textbf{65.0} & \textbf{62.3} & 54.8\\
    CSC\_LA\_SEM \cite{b8} & 66.5 & 64.4 & 61.2 & 53.1\\
    PointContrast\_LA\_SEM \cite{b7} & 65.3 & 63.6 & 61.4 & \textbf{55.0}\\
    \bottomrule
\end{tabular}
\label{tbl:benchmark}
\end{table}

\subsection{Viewpoint Bottleneck: Implementation}
After making aforementioned three simplifications to Eq.\ref{eq:ibf}, we obtain the final viewpoint bottleneck loss function as:
\begin{equation}
\mathcal{L}_{\rm VB}\triangleq \|\Gamma_{\lambda}(\mathcal{Z})-\mathcal{I}\|_F \label{eq:vb}
% \mathcal{L}_{\rm VB}\triangleq \sum_{i}\left(1-\mathcal{Z}_{ii}\right)^2+\lambda\sum_{i}\sum_{j\neq i}\mathcal{Z}_{ij}^2\label{eq:vb}
\end{equation}
where $\mathcal{I}$ denotes the identity matrix, $\Gamma_{\lambda}(\cdot)$ denotes scaling off-diagonal elements by positive constant $\lambda$, and $\mathcal{Z}$ is the cross-correlation matrix computed between the two representations of different viewpoints along the feature dimension:

\begin{equation}
\mathcal{Z} \triangleq \left(\widetilde{Z_p'}\right)^T\widetilde{Z_q'}\label{eq_cc}\\
% \mathcal{Z}_{ij}\triangleq\frac{\sum_{a}{Z_{a,i}'^pZ_{a,j}'^q}}{\|Z_{a,i}'^p\|_2\sqrt{\sum_{a}\left(Z_{a,j}'^q\right)^2}}\label{eq_cc}
\end{equation}

$\widetilde{Z'}$($H\times D$) is the representation obtained by normalizing $Z'$ along the sample dimension.  % with $a$ as the index inside a batch, $i,j$ as the vector dimension of the representations and $p,q$ indicating two different viewpoints. 

$\mathcal{L}_{\rm VB}$ aims to push the diagonal elements of $\mathcal{Z}$ towards 1, so that the representations are invariant to random geometric transformations. Meanwhile, by forcing the off-diagonal elements towards 0, we decorrelate different vector components of the representations, so that they contain non-redundant information about the original point cloud.

% Obviously, the first term of the loss function makes the representations invariant to the different viewpoints applied, by making the diagonal elements of the cross-correlation matrix to 1. The second term of the loss function decorrelates the different vector components of the representations, through making the off-diagonal elements of the cross-correlation matrix to 0. This decorrelation reduces the redundancy between the output representations, so that the representations include the non-redundant information about the origin point cloud of the scene.

\subsection{Pointly-supervised Scene Parsing}

After self-supervised representation learning with the proposed viewpoint bottleneck loss $\mathcal{L}_{\rm VB}$, we fine-tune on the pre-trained backbone by adding a scene parsing head (shown as light blue blocks in Fig.\ref{fig:overview}) with $\{P_i,S_i\}$. This training scheme is much better than directly training with $\{P_i,S_i\}$, since the intrinsic structure between enormous unlabelled points are fully leveraged by self-supervised pre-training.
%The major motivation of our SSRL pre-train method is that it can transfer well to weakly-supervised 3D semantic segmentation task. As shown in Fig.\ref{fig:overview}-(B), we fine-tune the pre-trained model on the ScanNetV2 dataset\cite{b3} which is denoted as $\{P_i, L_i\}$. $P_i (N\times6)$ is the 3D coordinates and the RGB values of points and $L_i\in\{0,1,2,3,..,40\}$ is the label map in which only 20, 50, 100 or 200 points have annotations. We adopt the Minkowski Convolutional Neural Network \cite{b34} as structured segmentation model, and obtain inference results on 20 categories ($\{1,2,3,4,5,6,7,8,9,10,11,12,14,16,24,28,33,34,36,39\}$). 

%The results are shown in Table.~\ref{tbl:table1}. After using the pre-training model, we obtain an $0.96$ improvement. This margin demonstrates the effectiveness of viewpoint bottleneck on semantic segmentation task.

%\begin{table}[t]
%\centering
%\caption{results on the ScanNet semantic segmentation benchmark}
%\begin{tabular}{lc}
%    \toprule
%       & mIOU\\
%    \midrule
%    Baseline &  70.98\\
%    Baseline + Viewpoint Bottleneck(256) & 71.94(+0.96)\\
%    \bottomrule
%\end{tabular}
%\label{tbl:table1}
%\end{table}

\section{Experiments}

%The following section describes our experiments on the ScanNetV2 dataset, on which we obtain state-of-the-art results of weakly-supervised semantic segmentation task. 
\subsection{Datasets}
\subsubsection{Dataset for SSRL}
In this study, we use the official training split of ScanNet for unsupervised representation pre-training. ScanNet is a large-scale RGB-D video dataset with 3D mesh reconstructions of indoor scenes. It contains over 1500 scans reconstructed from around 2.5 million views. This dataset is annotated with both point-level and instance-level semantic labels. These labels are defined in a protocol of 20 categories. The official training split has 1201 scans.  %Here, we generate a pair of different viewpoints on scenes of ScanNet for the pre-training framework shown in Fig.~\ref{fig:overview}.

\subsubsection{Dataset for pointly-supervised 3D scene parsing}
For fair comparison, we use the point annotations provided by \cite{b8} during fine-tuning. Four settings are evaluated, in which only 20 points, 50 points, 100 points, or 200 points of one point cloud in the training split have semantic labels. Other unlabeled points are ignored during the calculation of cross entropy loss Eq.~\ref{eqce}. We firstly evaluate on the official validation set of 312 scans, using the mean intersection of union (mIOU) metric. We also provide quantitative results on the held-out test set, which is reported by an online server\footnote{http://kaldir.vc.in.tum.de/scannet\_benchmark/data\_efficient/}. %We use four different training configurations including 20, 50, 100 or 200 labeled points of per scene in ScanNetV2. We use 1201 pointly-supervised scans for training and use ScanNetV2 validation set of 312 fully-supervised scenes for evaluating the model performance.

\subsection{Implementation details}

Then, we sample a combination of following transformations to apply to the selected point cloud:
\begin{itemize}
    \item[1] Random rotation. We rotate the point cloud along z-axis by a random angle sampled from $\mathcal{U}[0, 2\pi]$.
    \item[2] Random mirroring. For each axis, we apply mirroring transformation with a probability of 0.5.
    \item[3] Random chromatic jitter. For each channel, we apply a noise sampled from $\mathcal{N}(0, 255 \times 0.05)$.
\end{itemize}

Our Sparse ConvUnet is implemented with the MinkowskiNet codebase \cite{b34}, taking both 3D coordinates and point-wise RGB values as inputs. After farthest point sampling, a subset of 1024 points are selected as the abstraction of a point cloud. To investigate the robustness of VB, we inspect three settings for feature dimension $D$: 256, 512, and 1024.

For SSRL, all experiments are trained with a batch size 2 for 20000 iterations on two GeForce RTX 3090 GPUs, using a SGD optimizer with a momentum of 0.99 and an initial learning rate of 0.1. The learning rate is decayed according to a polynomial rule.

The pointly-supervised scene parsing experiments are trained with a batch size 12 for 30000 iterations on two GeForce RTX 3090 GPUs with the same optimizer setting. For both pre-training and fine-tuning experiments, the voxel size for Sparse ConvUNet is set to 2.0 cm. 

\subsection{Results and Analysis}

\subsubsection{Results on ScanNet validation set} Fristly, we show the effectiveness of viewpoint bottlenck on the ScanNet validation set. Quantitative results are summarized in Table.\ref{tbl:table2}. Baseline means training on point supervision without any SSRL. CSC stands for the contrsative scene context method \cite{b8}, which is a contrastive learning scheme applied on local point cloud partitions. Our method is denoted as VB with feature dimension $D$ marked in brackets. For fair comparison, the network architectures are exactly the same. 

Viewpoint bottleneck achieves clearly better results than CSC, in all four settings. In the 50 points setting, VB outperforms CSC by +3.2 mIOU. In the 20 and 100 points settings, the positive margins are also clear. In the 200 points setting, due to performance saturation, our quantitative results are comparable to CSC. Note that better performance in the down-stream pointly-supervised scene parsing task is not the only advantage of viewpoint bottleneck. As mentioned before, our method is easier to implement and tune than contrastive learning, as no sample pairs need to be selected.

Our method also out-performs the baseline by large margins. As shown in Table.~\ref{tbl:table2}, in the 20 points setting, VB brings +11.1 mIOU over the trivial point supervision baseline. This demonstrates that viewpoint bottleneck is an effective solution for ponitly-supervised 3D  scene parsing, as rich representations that capture the intrinsic structure of point clouds are firstly learnt in an unsupervised manner. An interesting fact is that with the supervision point number decreases, the margins grow bigger, for both CSC and VB.

Qualitative results with different numbers of supervised points are shown in Fig.\ref{fig:all_results}. With the number of supervision points increasing, from 20 points to 200 points, the parsing results are getting closer to the fully-supervised setting. More qualitative results and category-wise quantitative results are provided in the supplementary material. 

\subsubsection{Results on ScanNet test set}

We also evaluate on the held-out test set by submitting to an oneline server, with results shown in Table.~\ref{tbl:benchmark}. Only comparable 3D SSRL methods are listed. Compared with CSC\_LA\_SEM \cite{b8} and PointContrast\_LA\_SEM \cite{b7}, We achieve better performance in three settings. In the 20 points setting, our results are roughly the same as \cite{b7}. There are other complicated method with iterative pseudo label generation like \cite{b36}, which are not compared here due to the concern of fairness.

\subsubsection{Results with full supervision}

Although our goal is effective pointly-supervised scene parsing, we also evaluate with full supervision. As shown in Table.\ref{tbl:table1}, VB pre-training improves the fully supervised baseline by +0.9 mIOU.

\section{Conclusion}
In this paper, we address the challenging problem of learning 3D scene parsing models with only semantic annotations on sparse points. We develop a self-supervised representation learning framework named viewpoint bottleneck. We leverage a principled objective motivated by information theory and develops it into a practical surrogate loss. Pre-training with viewpoint bottleneck captures rich intrinsic structures between enormous unlabelled points. Pointly-supervised scene parsing performance on the public benchmark ScanNet is significantly improved, while reaching a new state-of-the-art among 3D SSRL methods.%we propose a SSRL pre-train method for pointly-supervised 3D Semantic Segmentation, named Viewpoint Bottleneck. Using this pre-train model, the results on ScanNet 3D Semantic label with Limited Annotations benchmark manifest that our approach outperforms the other SSRL methods on 3D semantic segmentation. We hope these findings will encourage more research on 3D SSRL.

\begin{table}[t]
\centering
\caption{Results on ScanNet validation set using full supervision}
\begin{tabular}{lc}
    \toprule
      & mIOU\\
    \midrule
    Baseline &  71.0\\
    VB (1024) & \textbf{71.9 (+0.9)}\\
    \bottomrule
\end{tabular}
\label{tbl:table1}
\end{table}

% \begin{table}[t]
% \centering
% \caption{Effect of FPS}
% \begin{tabular}{lcccc}
%     \toprule
%       & mIOU\\
%     \midrule
%     $w/\   FPS$ & 66.9 \\
%     $w/o\  FPS$ & 66.6 \\
%     \bottomrule
% \end{tabular}
% \label{tbl:FPS  (512) 100POINTS}
% \end{table}

%\begin{figure}
%    \centering
%    \includegraphics[width=0.4\textwidth]{loss.png}
%    \caption{Fine-tuning with Viewpoint Bottleneck(512) pre-trained model}
%    \label{fig:loss}
%\end{figure}

%%%%%%%%%%%%%%%%%%%%%%%%%%%%%%%%%%%%%%%%%%%%%%%%%%%%%%%%%%%%%%%%%%%%%%%%%%%%%%%%
%\section*{ACKNOWLEDGMENT}
%We are grateful to Anker Innovations for supporting this project.


\begin{figure*}[t]
\centerline{\includegraphics[width=1\textwidth]{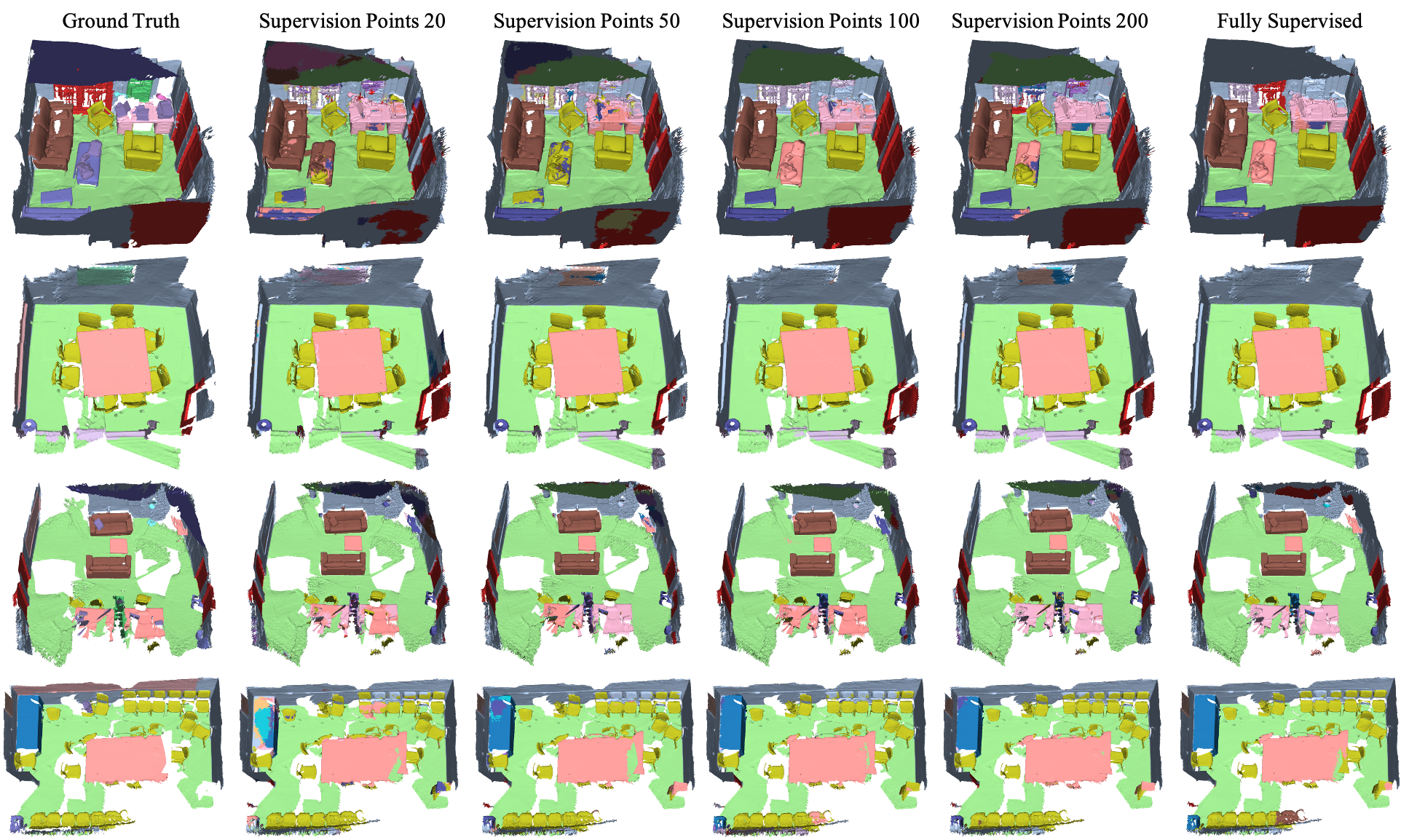}}
\caption{More qualitative results.}
\label{fig:all_results}
\end{figure*}

\addtocounter{figure}{-1}

\begin{figure*}[t]
\centerline{\includegraphics[width=1\textwidth]{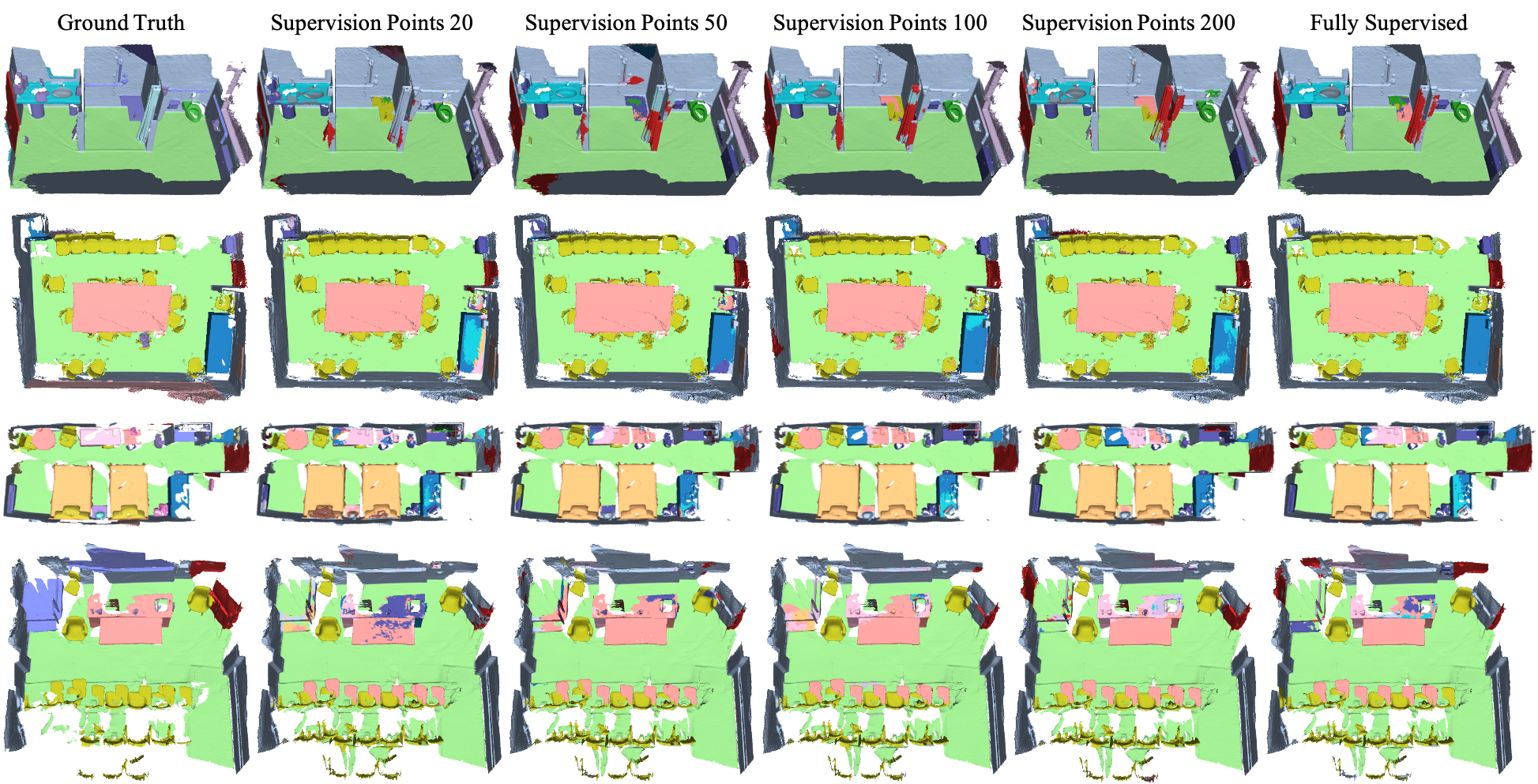}}
\caption{More qualitative results (cont.).}
\label{fig:all_results}
\end{figure*}

\addtocounter{figure}{-1}

\begin{figure*}[t]
\centerline{\includegraphics[width=1\textwidth]{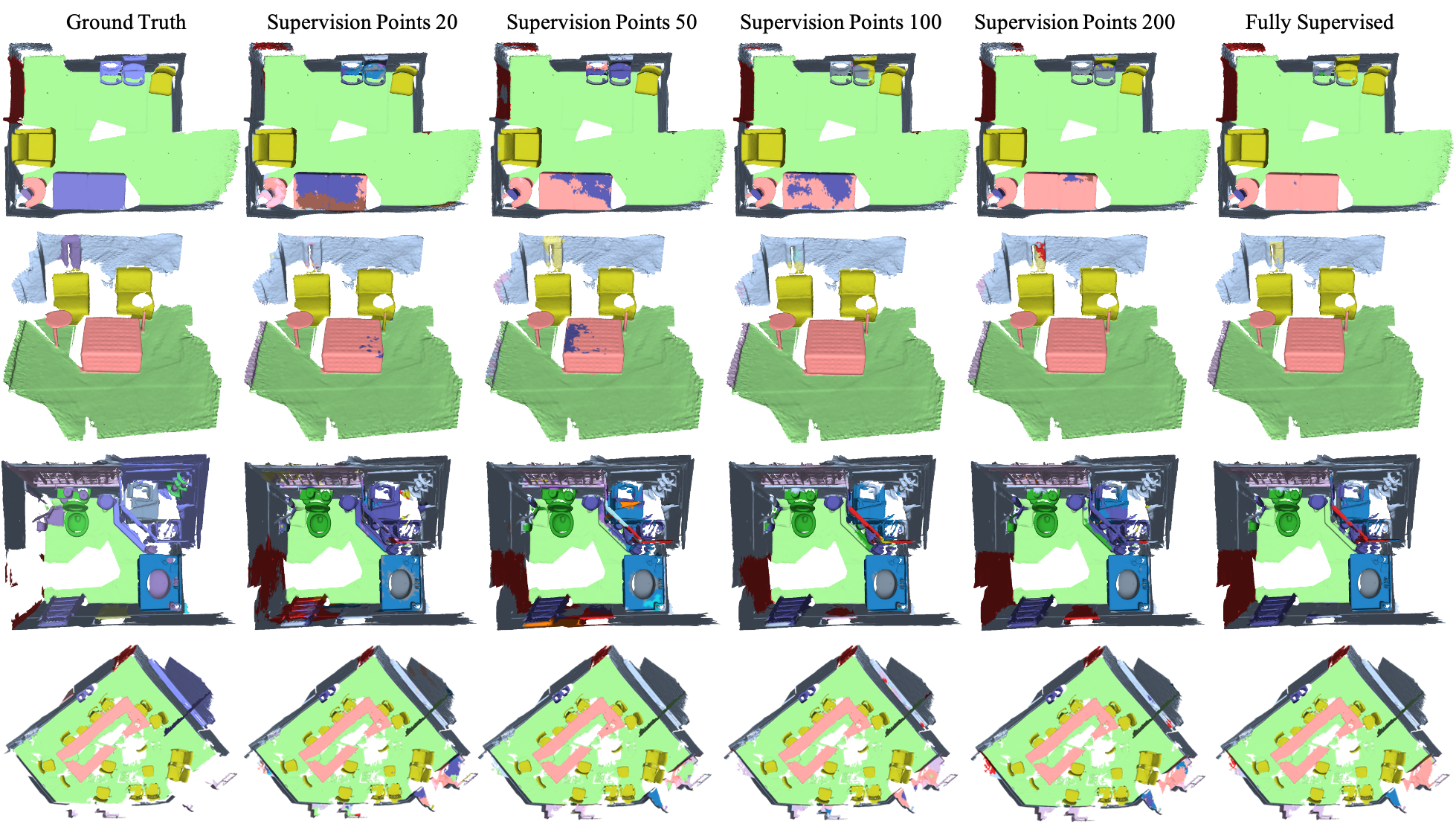}}
\caption{More qualitative results (cont.).}
\label{fig:all_results}
\end{figure*}

\addtocounter{figure}{-1}

\begin{figure*}[t]
\centerline{\includegraphics[width=1\textwidth]{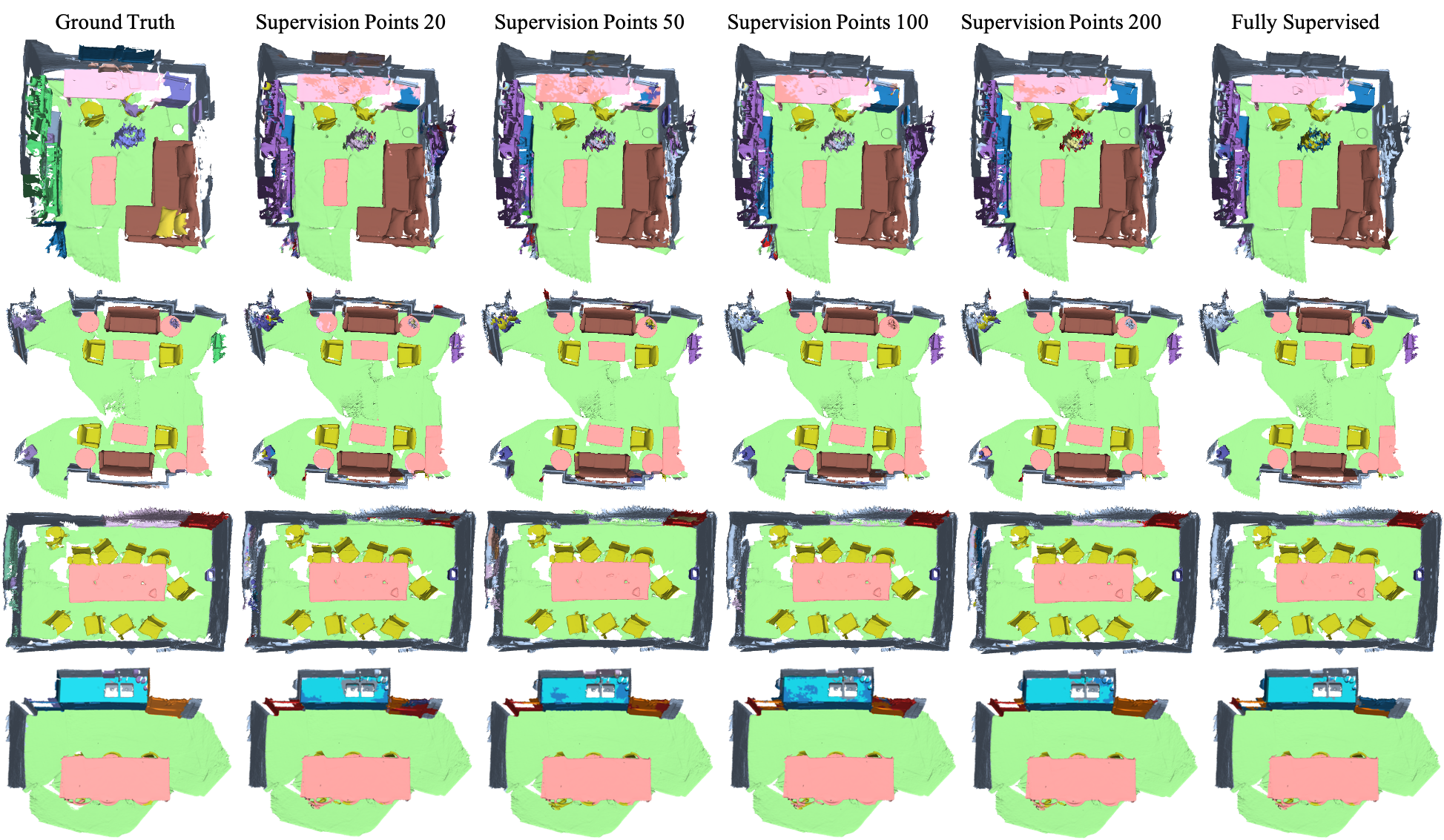}}
\caption{More qualitative results (cont.).}
\label{fig:all_results}
\end{figure*}

\addtocounter{figure}{-1}

\begin{figure*}[t]
\centerline{\includegraphics[width=1\textwidth]{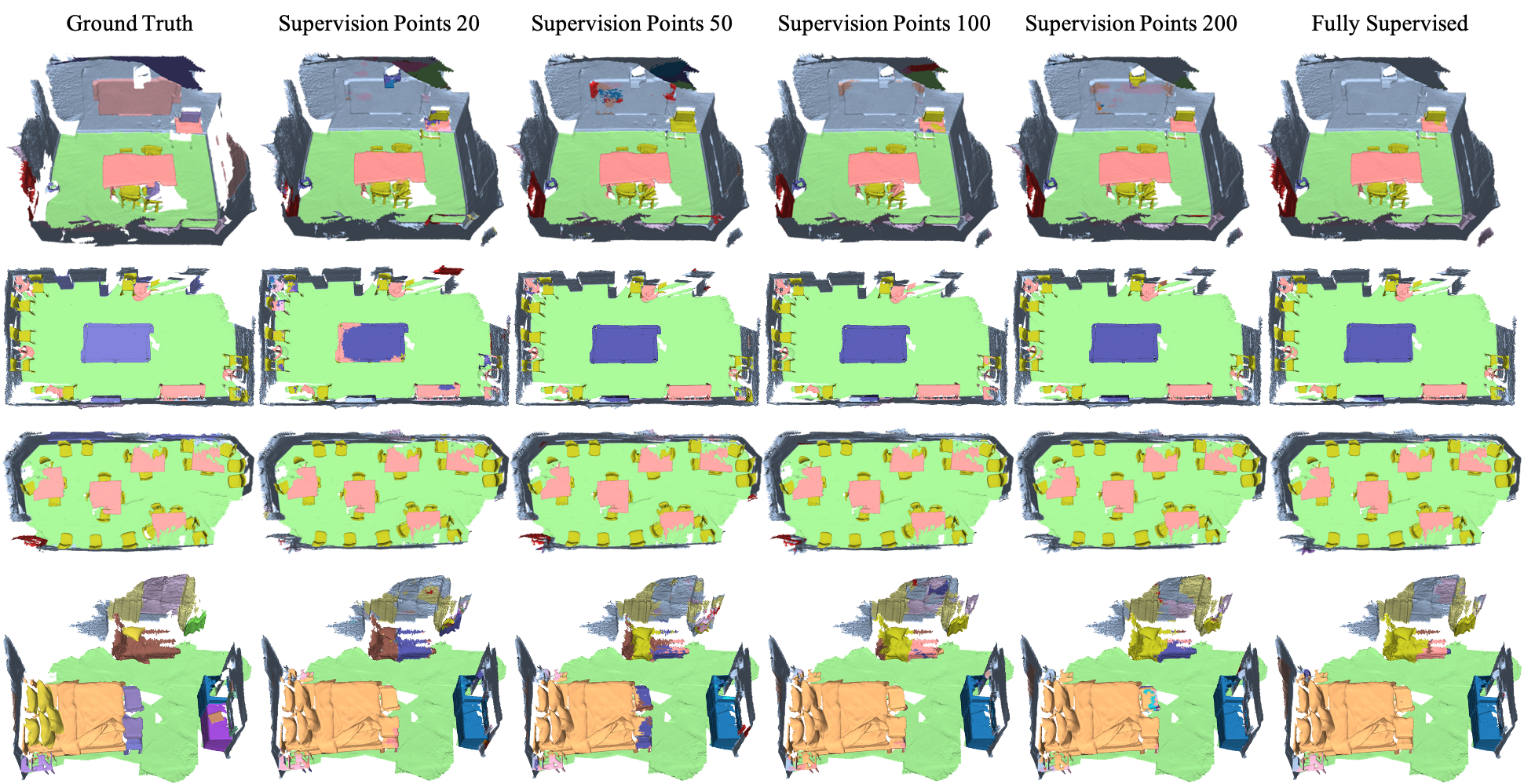}}
\caption{More qualitative results (cont.).}
\label{fig:all_results}
\end{figure*}

\addtocounter{figure}{-1}

\begin{figure*}[t]
\centerline{\includegraphics[width=1\textwidth]{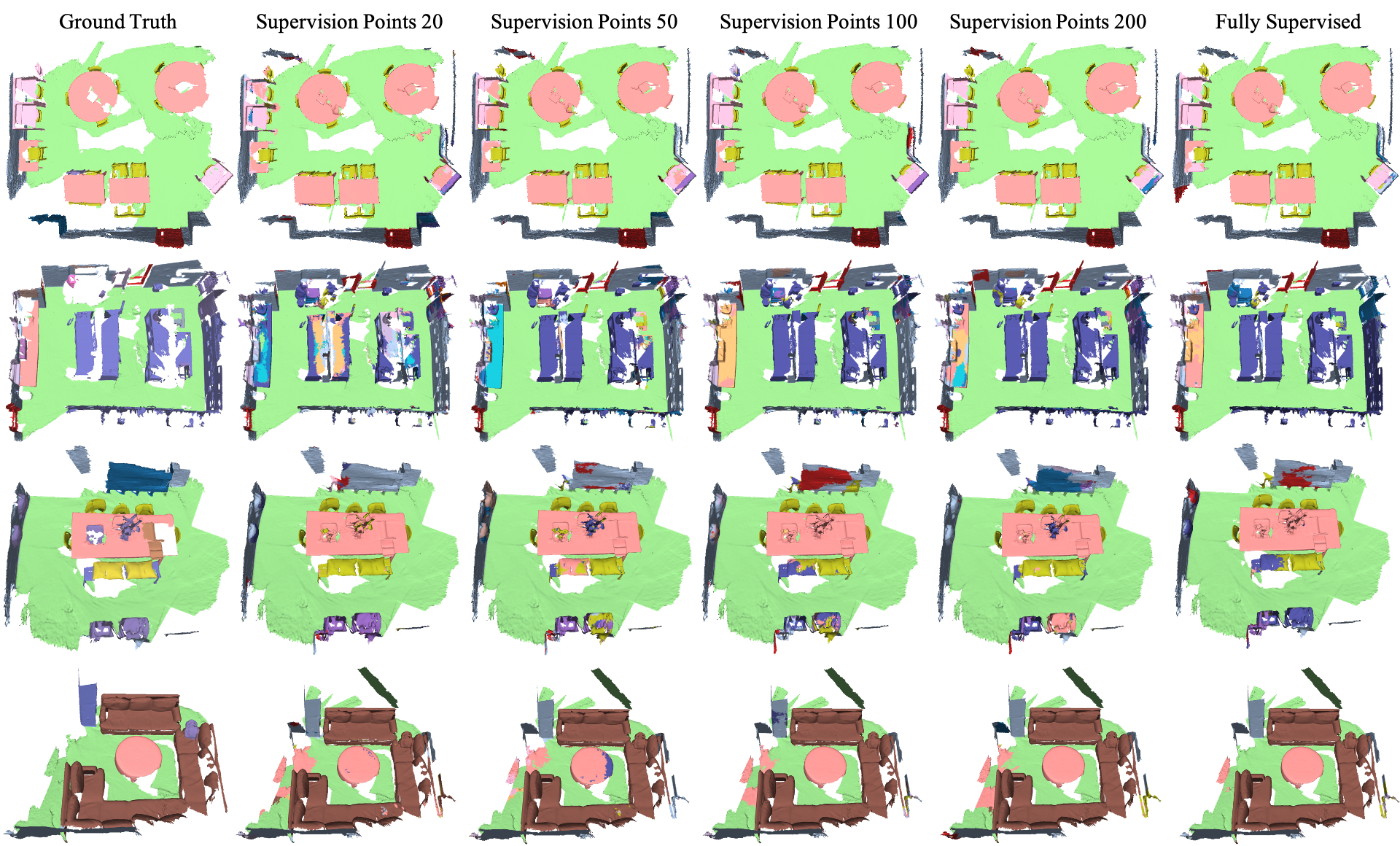}}
\caption{More qualitative results (cont.).}
\label{fig:all_results}
\end{figure*}

\begin{table*}
\centering
\caption{ScanNet 3D Semantic label with Limited Annotations Benchmark (mIOU:\%)}
\scalebox{0.95}{
\begin{threeparttable}
\begin{tabular}{cccccccccccccccccccccc}
    \toprule
    200 points & bathtub & bed & bookshelf & cabinet & chair & counter & curtain & desk & door & floor & AVG IOU\\
    \midrule
    % One-Thing-One-Click & 76.0  & 81.5 & 70.6 & 68.4 & 84.0 & 49.2 & 70.1 & 55.7 & 59.6 & 97.2 & 69.4\\
    Viewpoint\_BN\_LA\_AIR(ours) & 84.7 & 73.2 & 72.4 & 61.3 & 82.7 & \textbf{44.3} & \textbf{74.2} & \textbf{56.2} & 55.1 & 94.7 & \textbf{66.9}\\
    CSC\_LA\_SEM                 & \textbf{85.7} & 75.6 & \textbf{76.3} & \textbf{64.7} & \textbf{85.2} & 43.2 & 68.4 & 54.3 & 51.4 & \textbf{94.8} & 66.5\\
    PointConstrast\_LA\_SEM      & 71.7 & \textbf{77.5} & 75.4 & 62.6 & 80.4 & 39.1 & 68.9 & 48.5 & \textbf{57.2} & 94.5 & 65.3\\
    \midrule
    200 points & O.F. &  picture & refrigerator & S.C. & sink & sofa & table & toilet & wall & window\\
    \midrule
    % One-Thing-One-Click & 49.7 & 28.1 & 70.9 & 75.7 & 68.9 & 78.9 & 60.0 & 90.7 & 86.4 & 67.1 \\
    Viewpoint\_BN\_LA\_AIR(ours) & 44.1 & 21.8 & \textbf{65.0} & 75.3 & \textbf{62.1} & 76.5 & 60.1 & 90.5 & 81.4 & \textbf{61.8} \\
    CSC\_LA\_SEM & \textbf{46.9} & 17.9 & 59.9 & 70.2 & 62.0 & \textbf{78.9} & \textbf{61.4} & \textbf{91.1} & 81.5 & 60.7 \\
    PointConstrast\_LA\_SEM & 44.8 & \textbf{23.2} & 60.3 & \textbf{81.3} & 59.1 & 77.5 & 53.7 & 88.5 & \textbf{81.6} & 60.8\\
    \bottomrule
    \toprule
    100 points & bathtub & bed & bookshelf & cabinet & chair & counter & curtain & desk & door & floor & AVG IOU\\
    \midrule
    % One-Thing-One-Click & 73.4 & 81.5 & 66.1 & 64.4 & 84.1 & 50.9 & 74.1 & 47.9 & 54.8 & 96.8 & 67.0\\
    Viewpoint\_BN\_LA\_AIR(ours) & \textbf{77.8} & 73.1 & 68.8 & 61.7 & 81.2 & 44.6 & \textbf{73.9} & \textbf{61.8} & \textbf{54.0} & 94.5 & \textbf{65.0}\\
    CSC\_LA\_SEM & 76.1 & 70.7 & 70.3 & 64.2 & 81.3 & 43.6 & 65.9 & 50.2 & 51.6 & 94.5 & 64.4\\
    PointContrast\_LA\_SEM & 69.4 & \textbf{73.8} & \textbf{73.1} & \textbf{65.3} & \textbf{81.7} & \textbf{46.7} & 65.1 & 51.7 & 52.2 & \textbf{94.6} & 63.6\\
    \midrule
    100 points & O.F. &  picture & refrigerator & S.C. & sink & sofa & table & toilet & wall & window\\
    \midrule
    % One-Thing-One-Click & 46.1 & 25.1 & 64.4 & 75.4 & 65.6 & 74.4 & 54.1 & 91.7 & 84.4 & 62.5 \\
    Viewpoint\_BN\_LA\_AIR(ours) & 41.5 & 20.4 & \textbf{62.3} & 67.6 & 59.4 & 74.4 & \textbf{57.6} & 86.8 & \textbf{81.1} & 58.2 \\
    CSC\_LA\_SEM & \textbf{48.7} & \textbf{23.8} & 53.8 & \textbf{67.8} & \textbf{65.9} & 73.9 & 56.8 & \textbf{91.5} & \textbf{81.1} & 56.6 \\
    PointContrast\_LA\_SEM & 47.9 & 19.8 & 57.5 & 52.6 & 64.9 & \textbf{74.7} & 56.9 & 84.5 & 80.3 & \textbf{60.0} \\
    \bottomrule
    \toprule
    50 points  & bathtub & bed & bookshelf & cabinet & chair & counter & curtain & desk & door & floor & AVG IOU\\
    \midrule
    % One-Thing-One-Click & 72.5 & 73.5 & 71.7 & 63.5 & 82.9 & 45.7 & 63.9 & 42.1 & 55.2 & 96.7 & 64.2\\
    Viewpoint\_BN\_LA\_AIR(ours) & 81.2 & \textbf{74.3} & 65.4 & 57.9 & 80.0 & \textbf{46.2} & \textbf{71.3} & \textbf{53.3} & \textbf{51.6} & \textbf{94.4} & \textbf{62.3}\\
    CSC\_LA\_SEM & 74.7 & 73.1 & 67.9 & \textbf{60.3} & \textbf{81.5} & 40.0 & 64.8 & 45.3 & 48.1 & \textbf{94.4} & 61.2\\
    PointContrast\_LA\_SEM & \textbf{84.4} & 73.1 & \textbf{68.1} & 59.0 & 79.1 & 34.8 & 68.9 & 50.3 & 50.2 & 94.2 & 61.4\\
    \midrule
    50 points & O.F. &  picture & refrigerator & S.C. & sink & sofa & table & toilet & wall & window\\
    \midrule
    % One-Thing-One-Click & 46.0 & 24.0 & 55.8 & 78.8 & 62.1 & 72.0 & 47.7 & 91.5 & 84.2 & 53.9 \\
    Viewpoint\_BN\_LA\_AIR(ours) & \textbf{43.4} & \textbf{21.5} & 43.7 & 52.1 & \textbf{60.1} & \textbf{72.0} & \textbf{56.3} & \textbf{88.4} & \textbf{80.0} & 53.4 \\
    CSC\_LA\_SEM & 42.1 & 17.3 & \textbf{50.7} & 62.3 & 58.8 & 69.0 & 54.5 & 87.7 & 77.8 & \textbf{54.1} \\
    PointContrast\_LA\_SEM & 36.1 & 15.4 & 48.4 & \textbf{62.4} & 59.1 & 70.8 & 52.4 & 87.4 & 79.3 & 53.6\\
    \bottomrule
    \toprule
    20 points & bathtub & bed & bookshelf & cabinet & chair & counter & curtain & desk & door & floor & AVG IOU\\
    \midrule
    % One-Thing-One-Click & 75.6 & 72.2 & 49.4 & 54.6 & 79.5 & 37.1 & 72.5 & 55.9 & 48.8 & 95.7 & 59.4\\
    Viewpoint\_BN\_LA\_AIR(ours) & \textbf{74.7} & 57.4 & \textbf{63.1} & 45.6 & 76.2 & \textbf{35.5} & \textbf{63.9} & \textbf{41.2} & 40.4 & \textbf{94.0} & 54.8\\
    CSC\_LA\_SEM & 65.9 & 63.8 & 57.8 & 41.7 & 77.5 & 25.4 & 53.7 & 39.6 & \textbf{43.9} & 93.9 & 53.1\\
    PointContrast\_LA\_SEM & 73.5 & \textbf{67.6} & 60.1 & \textbf{47.5} & \textbf{79.4} & 28.8 & 62.1 & 37.8 & 43.0 & \textbf{94.0}& \textbf{55.0}\\
    \midrule
    20 points & O.F. &  picture & refrigerator & S.C. & sink & sofa & table & toilet & wall & window\\
    \midrule
    % One-Thing-One-Click & 36.7 & 26.1 & 54.7 & 57.5 & 22.5 & 67.1 & 54.3 & 90.4 & 82.6 & 55.7 \\
    Viewpoint\_BN\_LA\_AIR(ours) & \textbf{33.5} & \textbf{10.7} & 27.7 & \textbf{64.5} & 49.5 & 66.6 & \textbf{51.7} & 81.8 & 74.0 & 43.1 \\
    CSC\_LA\_SEM & 28.4 & 8.3 & \textbf{41.1} & 59.9 & 48.8 & \textbf{69.8} & 44.4 & 78.5 & 74.7 & 44.0\\
    PointContrast\_LA\_SEM & 30.3 & 8.9 & 37.9 & 58.0 & \textbf{53.1} & 68.9 & 42.2 & \textbf{85.2} & \textbf{75.8} & \textbf{46.8}\\
    \bottomrule
\end{tabular}
\begin{tablenotes}
\footnotesize
\item[*] The O.F. is the other furniture and the S.C. is the shower curtain.
\end{tablenotes}
\end{threeparttable}
}
\label{tbl:table3}
\end{table*}

\begin{table*}
\centering
\caption{Pointly-supervised 3D Parsing results on the ScanNet validation set (mIOU:\%)}
\scalebox{0.95}{
\begin{threeparttable}
\begin{tabular}{cccccccccccccccccccccc}
    \toprule
    200 points & wall & floor & cabinet & bed & chair & sofa & table & door & window & bookself & AVG IOU\\
    \midrule
    VB (256) & 82.3 & \textbf{95.6} & 60.8 & 78.6 & 88.8 & 82.2 & \textbf{72.4} & \textbf{59.6} & 52.9 & 76.1 & 68.4\\
    VB (512) & \textbf{82.8} & \textbf{95.6} & \textbf{61.3} & \textbf{81.6} & 89.0 & \textbf{84.6} & 70.0 & 59.1 & \textbf{55.0} & 75.2 & \textbf{68.5}\\
    VB (1024) &82.6 & 95.5 & 60.8 & 80.6 & \textbf{89.2} & 83.6 & 70.9 & 57.8 & 53.6 & \textbf{76.3} & 68.4\\
    \midrule
    200 points & picture &  counter & desk & curtain & refrigerator & S.C. & toilet & sink & bathtub & O.F.\\
    \midrule
    VB (256) & 27.5 & 59.2 & \textbf{60.8} & 68.3 & 48.3 & \textbf{66.9} & 87.8 & 63.4 & 86.6 & 49.2\\
    VB (512) & 27.1 & 59.5 & 58.5 & \textbf{69.3} & 44.2 & 63.1 & \textbf{89.9} & \textbf{63.6} & 85.6 & \textbf{55.0}\\
    VB (1024) &\textbf{29.9} & \textbf{60.1} & 59.4 & 68.6 & \textbf{49.1} & 61.9 & 89.7 & 61.7 & \textbf{87.1} & 49.8\\
    \bottomrule
    \toprule
    100 points & wall & floor & cabinet & bed & chair & sofa & table & door & window & bookself & AVG IOU\\
    \midrule
    VB (256) & 81.1 & 95.4 & \textbf{59.2} & \textbf{78.6} & 88.2 & 82.9 & 67.7 & 56.2 & 51.9 & 71.9 & 66.5\\
    VB (512) & \textbf{81.2} & \textbf{95.6} & 57.9 & 76.4 & \textbf{88.8} & \textbf{83.9} & \textbf{70.9} & \textbf{56.4} & \textbf{53.2} & 72.6 & \textbf{66.8}\\
    VB (1024) &81.0 & 95.5 & 58.6 & 77.7 & 88.7 & 82.6 & 70.2 & 54.5 & 51.0 & \textbf{74.0} & 66.5 \\
    \midrule
    100 points & picture &  counter & desk & curtain & refrigerator & S.C. & toilet & sink & bathtub & O.F.\\
    \midrule
    VB (256) & 23.9 & 55.1 & 55.0 & 65.4 & \textbf{50.8} & 64.0 & 89.7 & 57.9 & \textbf{85.5} & \textbf{48.6} \\
    VB (512) & 23.4 & \textbf{56.9} & 56.1 & \textbf{67.2} & 46.7 & \textbf{64.5} & 90.3 & \textbf{62.6} & 85.0 & 47.0\\
    VB (1024) &\textbf{24.9} & 55.9 & \textbf{56.5} & 65.3 & 47.1 & 59.2 & \textbf{92.6} & 60.5 & 85.1 & \textbf{48.6}\\
    \bottomrule
    \toprule
    50 points  & wall & floor & cabinet & bed & chair & sofa & table & door & window & bookself & AVG IOU\\
    \midrule
    VB (256) & 79.4 & \textbf{95.3} & 55.7 & \textbf{77.7} & 86.8 & 79.2 & \textbf{67.5} & 50.2 & 47.5 & \textbf{72.7} & 63.3\\
    VB (512) & \textbf{79.8} & 95.2 & 56.0 & 76.1 & \textbf{87.1} & 80.7 & 66.3 & \textbf{52.2} & \textbf{48.5} & 71.3 & 63.6\\
    VB (1024) &79.5 & 95.1 & \textbf{56.3} & \textbf{77.7} & \textbf{87.1} & \textbf{81.7} & 66.9 & 50.8 & 47.2 & 72.4 & \textbf{63.7} \\
    \midrule
    50 points & picture &  counter & desk & curtain & refrigerator & S.C. & toilet & sink & bathtub & O.F.\\
    \midrule
    VB (256) & \textbf{19.3} & \textbf{54.3} & \textbf{52.7} & 63.2 & 39.4 & 58.3 & 83.2 & \textbf{57.1} & 83.0 & 44.1\\
    VB (512) & 18.6 & 52.1 & 51.3 & \textbf{63.5} & \textbf{42.9} & 60.2 & 86.4 & 55.7 & 82.6 & \textbf{46.2}\\
    VB (1024) &15.5 & 51.1 & 52.0 & 62.7 & 40.3 & \textbf{62.5} & \textbf{89.2} & 56.2 & \textbf{84.1} & 44.8\\
    \bottomrule
    \toprule
    20 points & wall & floor & cabinet & bed & chair & sofa & table & door & window & bookself & AVG IOU\\
    \midrule
    VB (256) & 74.6 & 94.9 & 46.6 & 67.3 & 83.7 & 75.9 & 59.9 & 38.8 & 38.8 & 66.9 & 56.2\\
    VB (512) & 74.6 & \textbf{95.0} & \textbf{49.0} & \textbf{71.1} & \textbf{85.5} & \textbf{78.7} &\textbf{63.6} & \textbf{39.3} & 37.0 & 66.7 & \textbf{57.0} \\
    VB (1024) &\textbf{74.8} & 94.8 & 48.3 & 70.4 & 84.6 & 76.1 & 62.1 & 37.7 & \textbf{38.9} & \textbf{67.2} & 56.3\\
    \midrule
    20 points & picture &  counter & desk & curtain & refrigerator & S.C. & toilet & sink & bathtub & O.F.\\
    \midrule
    VB (256) & \textbf{14.3} & \textbf{45.6} & 44.1 & \textbf{55.9} & \textbf{30.7} & \textbf{52.5} & 76.3 & 46.0 & 78.1 & 33.8 \\
    VB (512) & 10.9 & 45.3 & \textbf{48.7} & 55.7 & 29.0 & 51.2 & \textbf{81.0} & \textbf{48.8} & 73.7 & 35.2 \\
    VB (1024) &11.5 & 43.9 & 43.5 & 54.0 & 30.1 & 46.0 & 78.8 & 48.2 & \textbf{78.6} & \textbf{36.3}\\
    \bottomrule
\end{tabular}
\begin{tablenotes}
\footnotesize
\item[*] The O.F. is the other furniture and the S.C. is the shower curtain.
\end{tablenotes}
\end{threeparttable}
}
\label{tbl:table3}
\end{table*}

\begin{table*}
\centering
\caption{Results on ScanNet validation set using full supervision (mIOU:\%)}
\scalebox{0.95}{
\begin{threeparttable}
\begin{tabular}{cccccccccccccccccccccc}
    \toprule
    Full Supervision & wall & floor & cabinet & bed & chair & sofa & table & door & window & bookself & AVG IOU\\
    \midrule
    Baseline & 84.9 & \textbf{96.0} & 63.1 & \textbf{81.6} & 90.1 & \textbf{83.6} & 73.1 & 62.8 & 59.1 & 80.0 & 71.0\\
    VB (1024) & \textbf{95.2} & \textbf{96.0} & \textbf{65.8} & 79.9 & \textbf{90.5} & 83.2 & \textbf{73.3} & \textbf{64.8} & \textbf{60.3} & \textbf{80.7} & \textbf{71.9}\\
    \midrule
    Full Supervision & picture &  counter & desk & curtain & refrigerator & S.C. & toilet & sink & bathtub & O.F.\\
    \midrule
    Baseline & \textbf{31.9} & 65.1 & 61.0 & \textbf{75.7} & 51.1 & \textbf{66.4} & 91.9 & 63.6 & \textbf{85.9} & 52.6 \\
    VB (1024) & 31.8 & \textbf{66.1} & \textbf{62.8} & 74.3 & \textbf{56.0} & 64.5 & \textbf{92.7} & \textbf{67.7} & 83.4 & \textbf{59.8}\\
    \bottomrule
\end{tabular}
\begin{tablenotes}
\footnotesize
\item[*] The O.F. is the other furniture and the S.C. is the shower curtain.
\end{tablenotes}
\end{threeparttable}
}
\label{tbl:table3}
\end{table*}

\end{document}